%% file: New_IEEEtran_how-to.tex
\documentclass[lettersize,journal]{IEEEtran}
\usepackage{amsmath,amsfonts}
\usepackage{algorithmic}
\usepackage{algorithm}
\usepackage{array}
\usepackage[caption=false,font=normalsize,labelfont=sf,textfont=sf]{subfig}
\usepackage{textcomp}
\usepackage{stfloats}
\usepackage{multirow}
\usepackage{multicol}
\usepackage{makecell}
\usepackage{setspace}
\usepackage{url}
\usepackage{svg}
\usepackage{verbatim}
\usepackage{graphicx}
\usepackage{hyperref}
\usepackage{cite}
\usepackage{xcolor}
\definecolor{citecolor}{HTML}{0071BC}
\hypersetup{colorlinks,linkcolor={red},citecolor={citecolor}}  
\def\BibTeX{{\rm B\kern-.05em{\sc i\kern-.025em b}\kern-.08em
    T\kern-.1667em\lower.7ex\hbox{E}\kern-.125emX}}
\usepackage{balance}
\newcommand{\my}[1]{\textcolor{black}{#1}}

\begin{document}
\title{RoadBEV: Road Surface Reconstruction in Bird's Eye View}



\author{
Tong Zhao$^{1}$ \quad Lei Yang$^{1}$ \quad Yichen Xie$^{2}$ \quad Mingyu Ding$^{2}$ \quad Masayoshi Tomizuka$^{2}$ \quad Yintao Wei$^{1}$\\
\hfill \break
$^{1}$Tsinghua University \quad $^{2}$UC Berkeley
}

\markboth{ 
}
{How to Use the IEEEtran \LaTeX \ Templates}

\maketitle

\input{sec/0_abstract}   


\input{sec/1_intro}

\input{sec/2_related}
\input{sec/3_dataset}

\input{sec/4_methods}
\input{sec/5_experiments}

\input{sec/limitation}
\input{sec/6_conslusion}

\bibliography{IEEEabrv}

\end{document}

%% file: sec/0_abstract.tex
\begin{abstract}
Road surface conditions, especially geometry profiles, enormously affect driving performance of autonomous vehicles. Vision-based online road reconstruction promisingly captures road information in advance. Existing solutions like monocular depth estimation and stereo matching suffer from modest performance. The recent technique of Bird's-Eye-View (BEV) perception provides immense potential to more reliable and accurate reconstruction. This paper uniformly proposes two simple yet effective models for road elevation reconstruction in BEV named RoadBEV-mono and RoadBEV-stereo, which estimate road elevation with monocular and stereo images, respectively. The former directly fits elevation values based on voxel features queried from image view, while the latter efficiently recognizes road elevation patterns based on BEV volume representing correlation between left and right voxel features. Insightful analyses reveal their consistence and difference with the perspective view. Experiments on real-world dataset verify the models' effectiveness and superiority. Elevation errors of RoadBEV-mono and RoadBEV-stereo achieve 1.83 cm and 0.50 cm, respectively. Our models are promising for practical road preview, providing essential information for promoting safety and comfort of autonomous vehicles.
The code is released at \href{https://github.com/ztsrxh/RoadBEV}{https://github.com/ztsrxh/RoadBEV}

\end{abstract}

%% file: sec/1_intro.tex
\section{Introduction}
\label{sec:intro}

\begin{figure*}[t]
  \centering
   \includegraphics[width=0.9\linewidth]{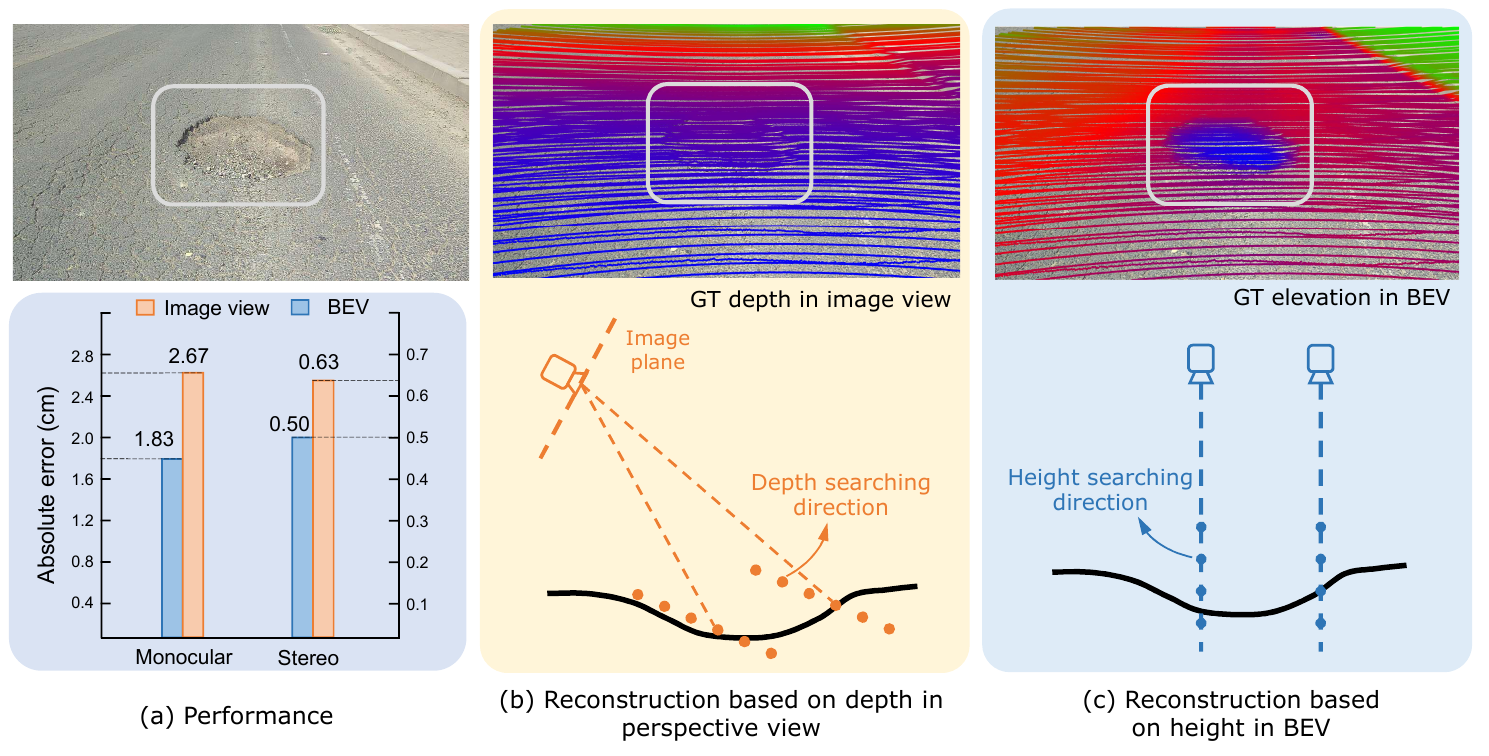}
   \vspace{-6pt}
   \caption{Our motivation. (a) Our reconstruction methods in BEV outperform these in image view for both monocular and stereo configurations. (b) For depth estimation in image view, the searching direction is biased from road elevation direction. Road profile features are sparse in depth view. The pothole is not clearly identifiable. (c) In BEV, profile vibrations are precisely captured such as the pothole, roadside step and even the rut. Road elevation feature in vertical direction is denser and easier to be recognized.}
   \label{fig:motivation}
\end{figure*}

\IEEEPARstart{I}{n} recent years, the rapid development of unmanned ground vehicles (UGVs) has posed higher requirements for on-board perception systems. Real-time understanding of driving environment and condition is vital for accurate motion planning and control \cite{8626459, ZHAO2024111019, 10367760}. For vehicles, road is the only contacting media with the physical world. Road surface conditions determine many vehicle characteristics and driving performance \cite{10101715}. Road unevenness like bumps and potholes (as shown in Fig. \ref{fig:motivation} (a)) exacerbates ride experience of passenger vehicles, which is intuitively perceptible \cite{10329453,ZHAO2022108483}. Real-time road surface condition perception, especially the geometry elevation, immensely contributes to planning and control systems in enhancing safety and ride comfort \cite{8324512, 9830854, LIANG2022109197}. 

Comparing with other perception tasks in UGVs like segmentation and detection, road surface reconstruction (RSR) is an emerging technique gaining more attention recently. Similar with existing perception pipelines, RSR generally utilizes on-vehicle LiDAR and camera sensors to retain road surface information. LiDAR straightforwardly scans road profile and derives point clouds \cite{ni2020road, weng2024big,9050489}. Although road elevation on tire trajectories can be directly extracted without complicated post-processing, LiDAR sensors cost much, limiting their applications on economical vehicles.

Image-based RSR, a 3D vision task, is more promising than LiDAR in terms of accuracy and resolution. Road surface texture is also preserved, enabling more comprehensive road perception. Road geometry reconstruction based on vision is actually a depth estimation task. For monocular camera, monocular depth estimation based on single image or multi-view stereo (MVS) based on image sequence can be implemented to directly estimate depth \cite{9691345}. For stereo camera, stereo matching regresses disparity map, which can be converted to depth \cite{6662410, 9025600,xin2024parameter}. Given camera parameters, road point clouds in camera's coordinate are recovered. Road structure and elevation information are finally obtained by primary post-processing pipelines. Supervised by ground-truth (GT) labels, high-accuracy and reliable RSR is achievable. 

However, RSR in image view suffers from inherent drawbacks. 
Depth estimation for a specific pixel is actually finding the optimal bins along the direction vertical to image plane, as indicated by the orange dots in Fig. \ref{fig:motivation} (b). The depth direction has a certain angle bias with road surface. The variation and trend of road profile features are inconsistent with that along the search direction. Informative clues about road elevation variation are sparse in depth view. Moreover, depth search ranges are the same for every pixel, leading models to capture global geometry hierarchy instead of local surface structure. Fine-grained road elevation is corrupted by the global but coarse depth searching. As we focus on elevation in the vertical direction, efforts made in the depth direction is wasted. Texture details at far distance are lost in the perspective view, which further poses challenges for effective depth regression without further prior constraints \cite{zhao2024depth}.

Estimating road elevation from top-down view (i.e., BEV) is a natural idea, as elevation inherently describes vertical vibration. BEV is an efficient paradigm for representing multi-modal and multi-view data in an uniform coordinate \cite{10321736}. Recent SOTA performance on 3D object detection and segmentation tasks are achieved by BEV-based methods \cite{10438483,xin2024vmt,chen2024taskclip}, which differs from the perspective view by introducing estimation heads on view-transformed image features. 
Fig. \ref{fig:motivation} shows our motivation. Instead of concentrating on global structure in image view, reconstruction in BEV directly recognize road features along vertical direction within a certain small range.
Projected road features in BEV densely reflects structure and profile variation, contributing to effective and fine-grained searching. The influence of perspective effect is also suppressed, as road is uniformly represented on a plane vertical to view angle. Road reconstruction based on BEV features is expected to achieve higher performance. 

In this paper, we reconstruct road surface in BEV to resolve the identified problems above. In particular, we focus on road geometry, i.e., elevation. To leverage monocular and stereo images and also, demonstrate the broad feasibility of BEV perception, we propose two sub-models named RoadBEV-mono and RoadBEV-stereo. Following the paradigm of BEV, voxels of interest covering potential road undulations are defined. The voxels query pixel features by 3D-2D projection. For RoadBEV-mono, elevation estimation head is introduced on reshaped voxel features. Structure of RoadBEV-stereo keeps consistent with the stereo matching models in image view. A 4D cost volume in BEV is built based on the left and right voxel features, which is further aggregated by 3D convolutions. Elevation regression is treated as classification on pre-defined bins for more efficient model learning. We validate the models on our previously released real-world dataset, showing their enormous superiority than the traditional monocular depth estimation and stereo matching methods.

Our contributions are summarized as follows:
\begin{itemize}
\item For the first time, we analytically and experimentally demonstrate the necessity and superiority of road surface reconstruction in BEV.

\item For both monocular and stereo-based schemes, we correspondingly propose two models named RoadBEV-mono and RoadBEV-stereo. Their mechanisms are explained in detail.

\item We comprehensively test and analyze performance of the proposed models, offering valuable insights and prospects for future research.
\end{itemize}

%% file: sec/2_related.tex
\section{Related Works}
\label{sec:related}

\noindent {\bf Road surface reconstruction by vision.}
Existing road reconstruction solutions are implemented in perspective view based on monocular or stereo images \cite{6957961}. Early works recover road profile and detect anomaly by introducing prior geometry constraints \cite{8500608, 8636338}. Based on known road surface disparity, road parallax or v-disparity map are built \cite{7797253}. Affine transformation based on v-disparity is performed to localize irregular unevenness \cite{feng2020road}. By introducing v-disparity road model and visual odometer, road elevation and drivable area are continuously extracted from stereo images. The geometry constraints above still rely on accurate disparity estimation. Road surface is sparsely reconstructed with structure from motion (SfM) based on sequence images in \cite{8794039}. They focus more on motion estimation with an adaptive Kalman Filter, which benefits continuous and global reconstruction in large-scale outdoor scenarios. The very recent research achieves large-scale monocular reconstruction via road mesh representations, recovering both geometry and texture  \cite{mei2024rome, wu2024emie}. However, road elevation suffers from poor accuracy as they focus more on texture while geometry are supervised by sparse labels.

\noindent {\bf BEV representation.} 
BEV representation offers a coherent perspective for autonomous driving, facilitating accurate object localization and streamlined fusion of data from multiple sensors~\cite{wang2023multi, song2024robustness, zhang2023dual,CAI2024102245}. This approach adeptly combines spatial and temporal data to enhance scene comprehension. Its applications span various real-world uses, including 3D object detection~\cite{song2024graphbev,10568349,yang2023bevheight, yang2024sgv3d, song2024robofusion}, occupancy prediction~\cite{huang2023tri}, motion planning~\cite{wang2022sti}, and online construction of HD maps~\cite{li2022hdmapnet}. Current works can be divided into two main categories based on view transformation: geometry-based and transformer-based. The transformer-based detectors, such as BEVFormer~\cite{li2022bevformer}, follow a methodology where they first design a collection of BEV grid queries. These queries are then utilized to facilitate the view transformation through cross-attention with image features. The geometry-based approaches, such as Lift-splat-shoot (LSS)~\cite{philion2020lift}, involve lifting each image into a frustum of features based on depth~\cite{huang2022bevdet4d} or height~\cite{yang2023bevheight, yang2023bevheight++} estimation and then splatting these frustums into a rasterized BEV grid. BEVDet~\cite{huang2021bevdet} directly projects images into the BEV space for 3D object detection. Subsequent research endeavors introduce depth supervision from LiDAR sensors~\cite{li2023bevdepth} or multi-view stereo techniques~\cite{li2022bevstereo} to enhance depth estimation accuracy, leading to cutting-edge performance levels in this domain. There is a natural consistency in spatial distribution between the road surface and BEV grids, which makes the BEV paradigm naturally suitable for RSR tasks. Building on the provision of horizontal information in BEV space, we further introduce height estimation, characterized by its dense distribution and ease of prediction by the network, to achieve accurate RSR.

%% file: sec/3_dataset.tex
\section{Dataset and Pre-processing} \label{sec:dataset}

We utilize our previously released dataset, \emph{Road Surface Reconstruction Dataset} (\emph{RSRD}) \cite{zhao2024road}, as a benchmark to test model performance. This is a large-scale and high-accuracy dataset special for road surface reconstruction purpose, providing 2,800 pairs of high-resolution stereo images, dense point cloud labels, as well as motion pose information in the dense subset. Unlike existing vision datasets for autonomous driving perception, it focuses only on road surface and retains rich road textures. It covers diverse conditions of asphalt and concrete roads, including typical even and  uneven cases like pothole and speed bump. We adopt the subset with half resolution (i.e., 960*540). For developing more reliable models and showcasing the significance of this application task, we extract representative samples with severer unevenness from the original dataset. The dataset utilized in this paper contains 1,210 training samples and 371 testing samples.

\begin{figure}[t]
  \centering
   \includegraphics[width=1\linewidth]{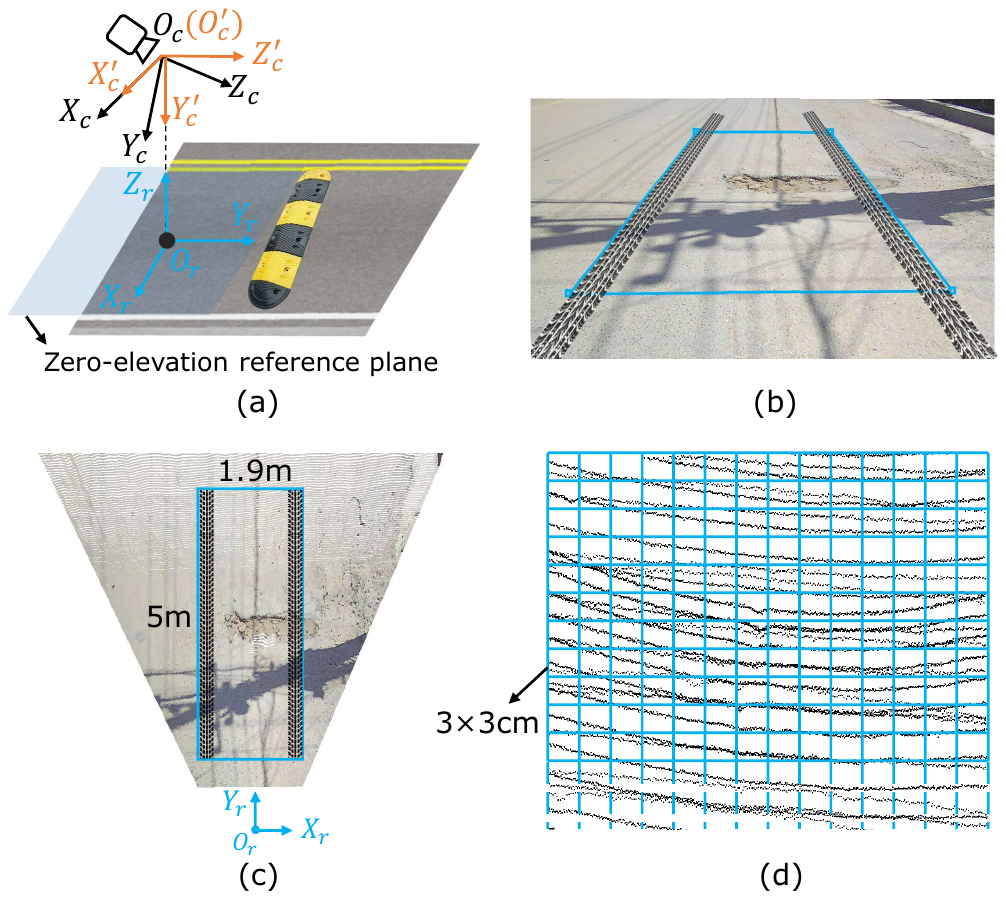}
   \vspace{-6pt}
   \caption{Illustration of coordinates and the generation of GT elevation labels. (a) Coordinates. (b) ROI in image view. (c) ROI in BEV. (d) Generation of GT labels in grids.}
   \label{fig:coord}
\end{figure}

We first introduce the coordinate definitions shown in Fig. \ref{fig:coord} (a). The $X_c\text{-}Y_c\text{-}Z_c$ is the original camera coordinate with a certain pitch angle w.r.t. the horizontal plane. The $X^{\prime}_{c}\text{-}Y^{\prime}_{c}\text{-}Z^{\prime}_{c}$ is an horizontal reference coordinate with  $X^{\prime}_{c}\text{-}Z^{\prime}_{c}$ in the horizontal plane and $Y^{\prime}_{c}$ pointing the vertical direction  (i.e., zero roll and pitch angles). The original and reference coordinates can transform to each other with the pose measured by IMU. Like depth estimation in image view where the camera plane indicates zero reference, road elevation also requires a reference base to describe profile properly. To facilitate algorithm development and the subsequent applications, we introduce another road coordinate $X_r\text{-}Y_r\text{-}Z_r$ vertically below the camera reference coordinate. The $X_r$ and $Y_r$ axes, which are parallel to $X^{\prime}_{c}$ and $Z^{\prime}_{c}$ respectively, represent the road lateral and longitudinal directions. The axis $Z_r$ now describes road profile, producing positive elevation values for roads above the reference plane; otherwise, negative. Based on our statistic analysis on the dataset, we set the referent height between $O^{\prime}_{c}$ and $O_r$ (i.e., the vertical distance between camera and road reference planes) as 1.10 m. Although the exact camera-road distance changes due to vehicle suspension compression and elongation, the variation is small and the feature voxel defined below can cover the slight change.

As we target to reconstruct road surface from a top-down perspective, view transformation is required to generate GT road elevation labels in BEV. Since only the road areas that vehicle passes through affect vehicle response, we focus on a certain ROI rather than the whole image. As illustrated by Fig. \ref{fig:coord} (b) and (c), we set the ranges along $X_r$ and $Y_r$ axes as [-1.0m, 0.9m] and [2.1m, 7.1m], respectively. The lateral range of 1.9 m covers width of most passenger cars, ensuring available road information on the left and right tire trajectories. The rectangular road area should be discretized to facilitate digital road elevation map. As shown in Fig. \ref{fig:coord} (d), we set the road grid resolution along both lateral and longitudinal directions as 3.0 cm, which is fine enough as the minimum road unevenness wavelength of interest is about 10 cm in automotive engineering. We obtain $Ny$ longitudinal and $Nx$ lateral grids, which are 164 and 64 in our settings.

\begin{figure}[t]
  \centering
   \includegraphics[width=1\linewidth]{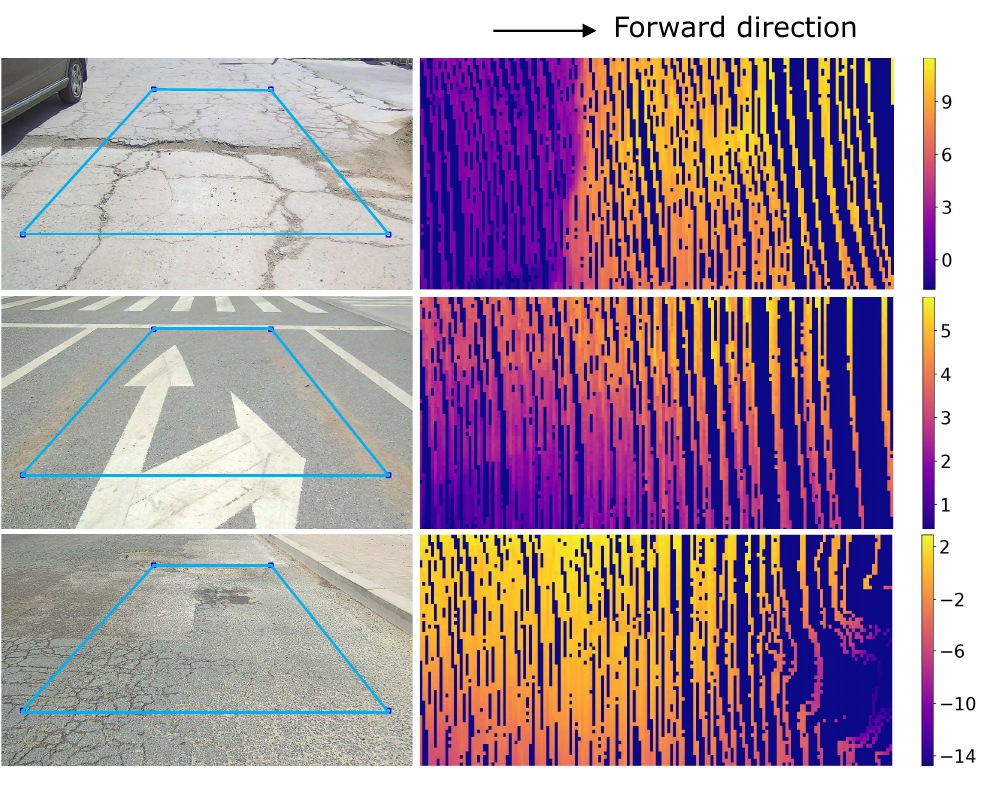}
   \vspace{-6pt}
   \caption{Examples of road image and GT elevation map. The unit of colorbar is cm.}
   \label{fig:map_example}
\end{figure}

The complete road surface point clouds in camera's coordinate are first transformed into camera reference coordinate, and then, the road reference coordinate. The points in ROI are cropped out. Points within every square grid of 3*3 cm size are then indexed and grouped. The GT elevation values of a grid is the average $Z_r$ coordinate value of its inner points. Since there may grids without any points falling in, a binary mask $\boldsymbol{M}$ is built to record the grids with available labels. GT elevation map $\boldsymbol{E}$ with shape 164*64 for every image sample is finally generated, as illustrated in Fig. \ref{fig:map_example}.

%% file: sec/4_methods.tex
\section{Methods} \label{sec:methods}
In this section, we first introduce general settings about feature voxel, view transformation, and elevation regression for the proposed models. The detailed structures of RoadBEV-mono and RoadBEV-stereo are then described.

\subsection{Feature Voxel and Elevation Regression} \label{feature voxel}
In the paradigm of BEV and occupancy perception, 3D voxels are first defined to facilitate the transformation and feature projection from perspective view to 3D view. For the mainstream detection or segmentation tasks, feature voxels generally cover large range at hundred meters with big intervals in the three dimensions. Nevertheless, for this RSR task whose scale is much smaller, voxel interval should also be reduced to ensure high accuracy. Balancing accuracy and computation, we set the vertical voxel interval as 1.0 cm, which is smaller than the lateral and longitudinal intervals of 3.0 cm. This resolution covers the amplitude of slight road fluctuations like cracks and small rocks. 

For object detection, the vertical range of interest is usually set as several meters above road, covering the height of most potential obstacles. The settings for road reconstruction are distinct as we focus road itself and also, road surface may be above or below the reference plane. Considering the practical road unevenness patterns, we set the elevation range as [-20cm, 20cm], resulting in $N_z$=40 voxels in the vertical direction. This elevation range covers the maximum extent of most common road unevenness like bumps and potholes. As mentioned in Section \ref{sec:dataset}, the actual height of camera relative to road is variable and road surface may be holistically above or below the reference plane. Despite that, the determined elevation range can still capture the trend and structure of roads. Finally, we obtain the feature voxel with shape 164, 64, and 40 on the $Y_r$, $X_r$, and $Z_r$ axes, respectively. 

\begin{figure}[t]
  \centering
   \includegraphics[width=1\linewidth]{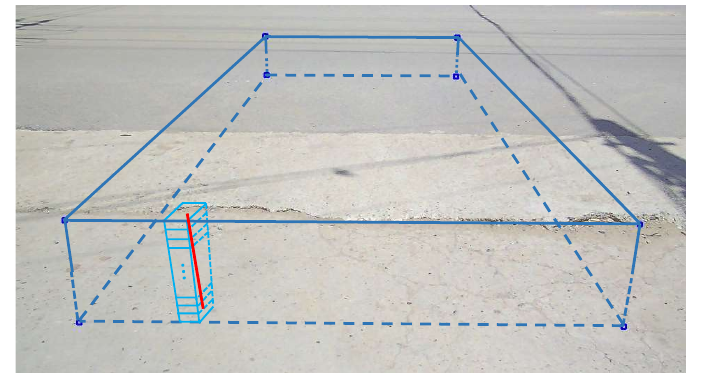}
   \vspace{-6pt}
   \caption{Feature voxels of interest in image view. Centers of stacked voxels at the same horizontal location are projected as pixels on the red line segment.}
   \label{fig:voxel}
\end{figure}

To achieve view transformation, we adopt 3D to 2D projection instead of 2D to 3D as we concentrate only on the specific ROI. Querying features of voxels by 3D to 2D is easier to implement. Another reason is that 2D to 3D requires accurate depth estimation, while we inherently aim to recover road surface geometry. For filling features to the voxels, we project the voxel centers to image plane with extrinsic and intrinsic parameters and index the corresponding pixel features. We visualize the feature voxels in image view, as shown in Fig. \ref{fig:voxel}. The pixel projections of stacked voxels at the same horizontal location are connected as a line segment. Voxels at further distance correspond to a shorter line segment. 

The target of our task is estimating continuous elevation values for every grid, which can be naturally defined as a regression task. However, end-to-end regression in deep learning often suffers from poor performance, as the search space is enormous. The commonly adopted loss functions for regression, such as L1 and MSE loss, cannot effectively constraint the learning of models. Therefore, we regard it as classification on pre-defined bins in the elevation range \cite{yang2023bevheight}. The GT values are also converted to one-hot labels on the corresponding bins. More details are given in Section \ref{Loss Functions}.

\subsection{RoadBEV-mono} \label{RoadBEV-mono}
\begin{figure*}[t]
  \centering
   \includegraphics[width=1\linewidth]{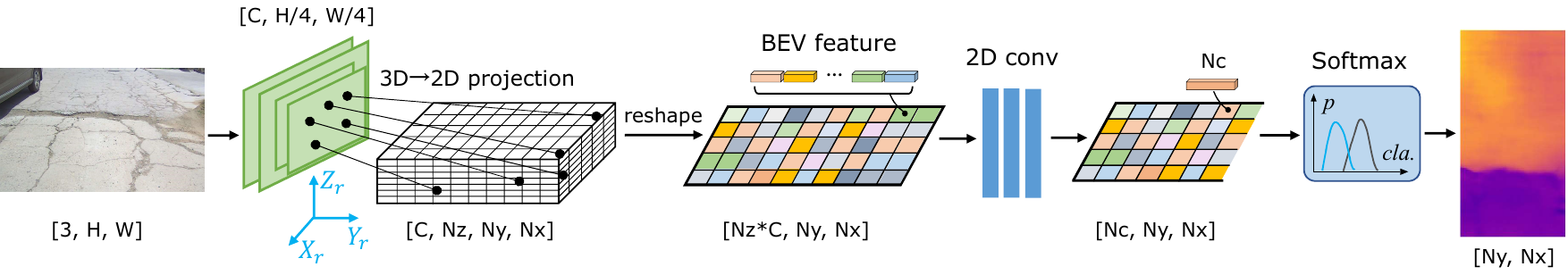}
   \vspace{-6pt}
   \caption{Architecture of RoadBEV-mono. We utilize 3D to 2D projection to query pixel features. The elevation estimation head utilizes 2D convolution to extract features on the reshaped BEV feature.}
   \label{fig:mono}
\end{figure*}

Fig. \ref{fig:mono} shows the architecture of RoadBEV-mono. The input RGB image undergoes a feature extraction backbone simplified from EfficientNet-B6 \cite{tan2019efficientnet}. Similar with the common structures in detection, it comprises a feature pyramid with $\frac{1}{{2^n}}$ resolutions, where $n \in$ \{1,2,3,4\}. The multi-scale feature maps retain both low-level geometry and high-level semantic information. The feature pyramid is then interpolated to $\frac{1}{4}$ resolution, concatenated along the channel dimension, and finally fused as maps with $C$=64 channels. The corresponding indexing pixels of feature voxels are determined by projecting coordinates of voxel centers to the image plane. The voxels are then filled with the $C$-dimensional pixel features, resulting in voxel feature $\boldsymbol{F}_{vox}\in \mathbb{R}^{C\times N_z\times N_y\times N_x}$. As demonstrated in \cite{xie2022m}, to facilitate more efficient feature extraction for downstream tasks, the BEV feature $\boldsymbol{F}_{BEV}\in \mathbb{R}^{(C\cdot N_z)\times N_y\times N_x}$ is derived by reshaping the vertical dimension of voxel feature. The $\boldsymbol{F}_{BEV}$ then undergoes 2D convolution with a reduced EfficientNet-B0, which is more cost-effective than the 3D.

Since we regress road elevation by classification on bins, a elevation feature map $\boldsymbol{F}_{ele}\in \mathbb{R}^{N_c\times N_y\times N_x}$ is generated, where $N_c$ is the number of classes (i.e., number of elevation bins). The channels of every grid are then normalized by \emph{softmax}, representing the possibilities of corresponding elevation values. The continuous elevation map  $\boldsymbol{\hat{E}}$ is predicted by utilizing the \emph{soft argmin} operation:

\begin{equation}
    \boldsymbol{\hat{E}}(y,x)=\sum_{c=1}^{N_c}{e_c\cdot Softmax({{\boldsymbol{F}_{ele}}(\cdot,\ y,\ x)})}
  \label{eq:ele_pred}
\end{equation}

\noindent where $e_c$ is the elevation value of $c\text{-}th$ bins, $y$ and $x$ denote the longitudinal and lateral dimension of the ROI grids, respectively. 

\begin{figure}[t]
  \centering
   \includegraphics[width=1\linewidth]{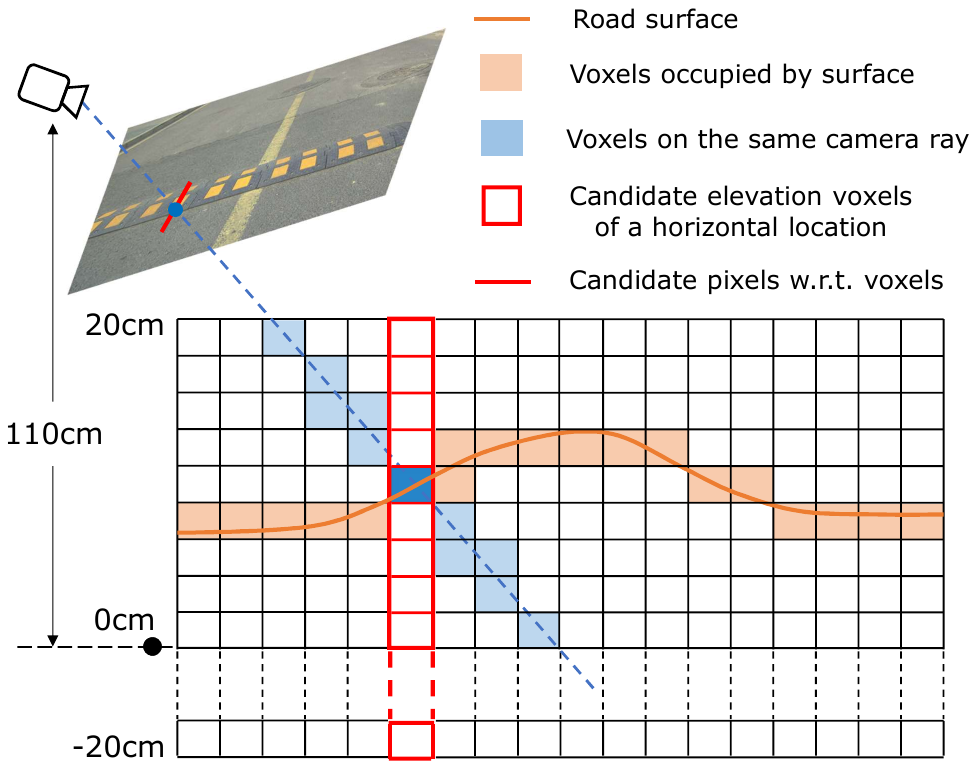}
   \vspace{-6pt}
   \caption{Mechanism of RoadBEV-mono. The voxels are in side view. }
   \label{fig:mono_mechanism}
\end{figure}

As illustrated in Fig. \ref{fig:mono_mechanism}, we further give insightful analysis on the mechanism of monocular-based RSR in BEV. The road surface, indicating a bump, falls in the orange voxels. The candidate voxels marked in red at a horizontal location are projected as pixels on the red line segment in image. Every voxel is a proposal querying image features that may be supporting it. However, there is generally only one elevation value at a horizontal location. Therefore, only one voxel among the candidates (as the deep blue voxel shows) contains the positive features, while the others receive negative features. The mission of the classification head is actually recognizing and emphasizing the correct feature while suppressing the others. The light blue voxels share the same pixel feature as they are on the same camera ray, indicating that positive features for current location may be negative for other locations. 

The mechanism analyzed above is actually the same with monocular depth estimation adopting search bins such as AdaBins \cite{9578024}. Our RoadBEV differs from them as the search is directly along height direction instead of depth. It directly recognizes feature patterns of elevation variation, avoiding ineffective efforts in a biased direction. The BEV paradigm for RSR is thereby expected to be more effective than depth. 

\subsection{RoadBEV-stereo} \label{RoadBEV-stereo}
\begin{figure*}[t]
  \centering
   \includegraphics[width=1\linewidth]{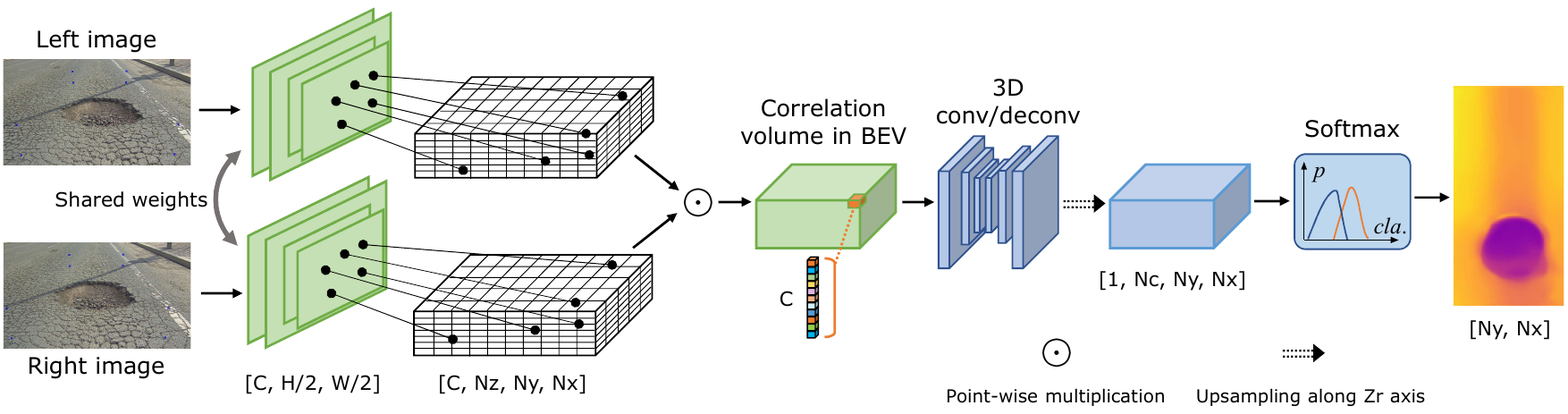}
   \vspace{-6pt}
   \caption{Architecture of RoadBEV-stereo. The voxel defined under the left camera's coordinate queries pixel features of the left and right feature maps. We construct cost volume in BEV by correlation between left and right voxel features. 3D convolutions then aggregate the 4D volume, predicting the occupancy possibility among vertical voxels.}
   \label{fig:stereo}
\end{figure*}

Fig. \ref{fig:stereo} shows the architecture of RoadBEV-stereo. Similar with RoadBEV-mono and general stereo matching models, the left and right images first undergo feature extraction with shared weights. Settings about the CNN backbone are the same as RoadBEV-mono, i.e., EfficientNet-B6. The only difference is that the feature map resolution is interpolated to $\frac{1}{2}$ in the FPN. Explanations and analysis about this resolution setting will be given below. The feature voxels are projected onto both the two image planes and query features from left and right perspectives. The left and right voxel features, $\boldsymbol{F}^{\prime}_{vox,l}$ and $\boldsymbol{F}^{\prime}_{vox,r}$, with the same shape $[C, N_z, N_y, N_x]$ are thereby obtained. In the architecture of stereo matching models, cost volume encoding similarities of the two feature maps is then built by correlation operation. We follow this paradigm and the voxel cost volume $\boldsymbol{V}_{discr}$ in BEV is derived as:

\begin{equation}
    \boldsymbol{V}_{discr}=\boldsymbol{F}^{\prime}_{vox,l} * \boldsymbol{F}^{\prime}_{vox,r}
  \label{eq:dis_voxel_feature}
\end{equation}

\noindent where * is the point-wise multiplication. The volume construction here differs from stereo matching in terms of feature inputs. The vertical dimension of the feature voxel represents elevation candidates or proposals, which is equivalent to the disparity dimension in stereo matching. For stereo matching, input features are 3D and the disparity dimension is introduced during volume construction. Instead, our proposed method directly takes 4D voxel features as inputs whose proposal dimensions are introduced before volume construction. 

Similar with RoadBEV-mono, the 4D BEV volume can be reshaped as a 3D tensor and processed with 2D convolutions, which efficiently saves computation. Nevertheless, to demonstrate its consistence with stereo matching models, we keep it as a 4D volume and aggregate by 3D convolutions. In this way, the estimation head is actually an occupancy network, which predicts the possibility of being occupied by road surface. Specifically, we stack six 3D-conv layers and three hourglass modules with $\frac{1}{2}$ resolution reduction. The channel dimension of BEV volume is finally reduced to one. Since the number of vertical voxels $N_z$ may be different from the number of elevation classes $N_c$, the elevation proposal dimension is interpolated to have $N_c$ channels. The elevation feature map $\boldsymbol{F}^{\prime}_{ele}\in \mathbb{R}^{N_c\times N_y\times N_x}$ is thereby derived. The following output layers are the same with RoadBEV-mono. \my{Different from general occupancy networks for scene perception where possibility is given for every voxel, the possibility here is normalized along the vertical dimension, indicating occupancy among elevation candidates at a horizontal grid.} 

\begin{figure}[t]
  \centering
   \includegraphics[width=1\linewidth]{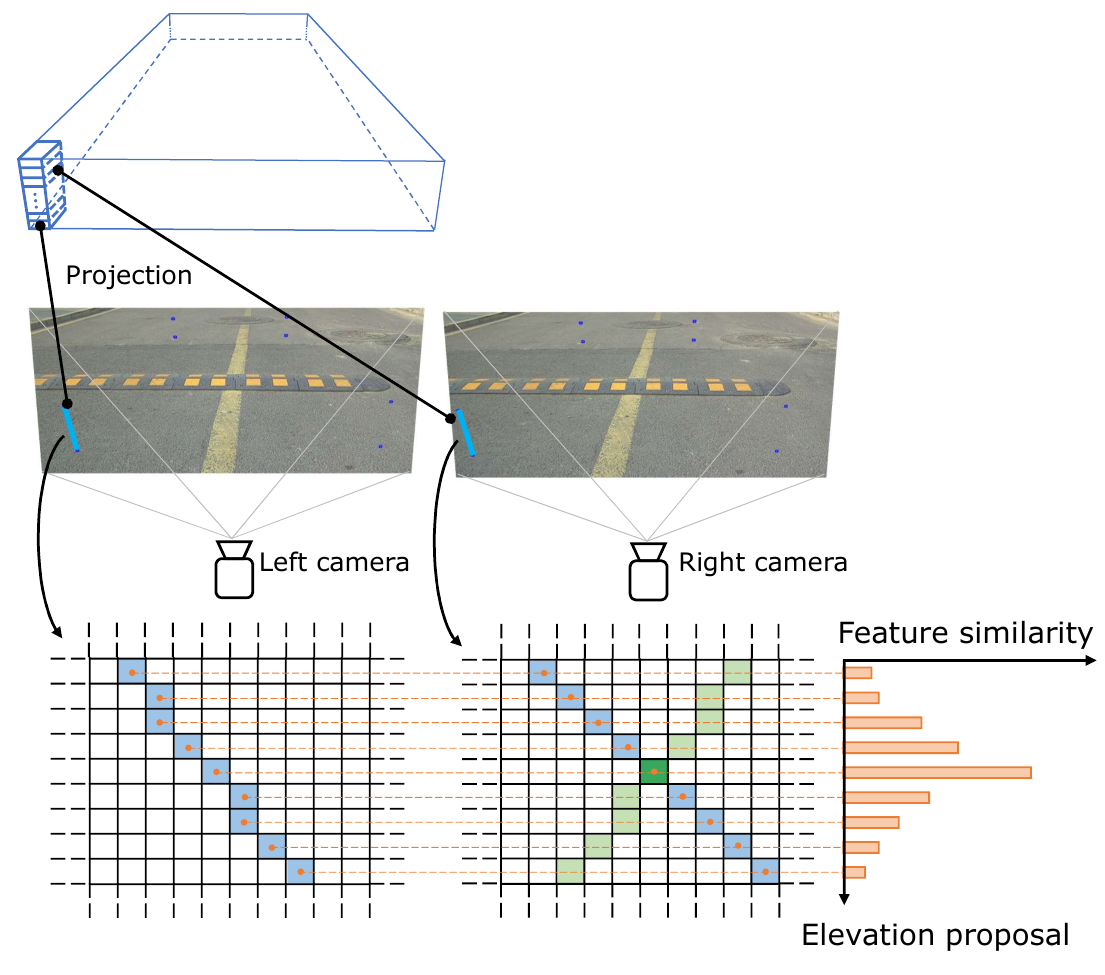}
   \vspace{-6pt}
   \caption{Mechanism of RoadBEV-stereo.}
   \label{fig:stereo_mechanism}
\end{figure}

Also, we further give insightful analysis on the mechanism of stereo-based RSR in BEV, as illustrated in Fig. \ref{fig:stereo_mechanism}. The staked voxel proposals at a horizontal location are projected onto both the left and right image planes, as denoted by the blue pixels. The right voxel features provide information of same voxel locations from a new view. The green pixels refer to the same real-world positions with blue pixels in left image, i.e., correspondence points. They correspondingly have high feature similarity. The green line segment indicating correspondence pixels will have only one intersection with the blue line segment indicating querying pixels. The queried feature pair has low similarity for an incorrect elevation proposal, as the correspondence pixel is away from query pixel in right image. The actual elevation location, which is projected as the deep green pixel, has highest similarity or lowest cost among all the pairs.  

For stereo matching in perspective view, the mission is finding the correspondence with highest feature similarity among pixels on the same row of right image. The search range is the width of image. For our method, the mission of classification head is identifying the highest similarity among proposal pairs. The search range is the number of elevation candidates, which is much smaller than image width. Theoretically, the estimation mechanisms are the same, i.e., measuring feature similarities and finding the highest. Nevertheless, the search range of stereo matching in BEV enormously narrows down than that in perspective view. The shrinking of candidate range contributes to effective aggregation by considering less unrelated feature measurements.

Despite the benefits, requirements for accurate RSR with stereo images in BEV are more strict. High-resolution feature map is necessary to cooperate with the precise but narrow search range. Road image features are similar in a certain area as road texture are repetitive without regular patterns. High-resolution feature is therefore vital for identifying slight feature variation. This is also the reason why we leverage image feature maps with $\frac{1}{2}$ resolution. 

\subsection{Loss Functions} \label{Loss Functions}
As demonstrated above, we regress road elevation by classification on bins. We set the class interval as 0.5 cm, indicating $N_c$=80 for the elevation range of [-20cm, 20cm]. The reason of this setting is explained in Section \ref{ablation_mono}. The total loss for an elevation map predicted by RoadBEV-mono is the sum of cross entropy losses for every grid $g$ with valid label:

\begin{equation}   
\mathcal{L}=-\sum_{g}^{}{\boldsymbol{M}(g)\cdot \sum_{c}^{N_c}{\boldsymbol{E}(c,g)\cdot log(Softmax({{\boldsymbol{F}_{ele}}(\cdot,g)}))}}
  \label{eq:total_loss}
\end{equation}

\noindent where $\boldsymbol{M}$ is the binary GT mask, $c$ is the class index. The loss for RoadBEV-stereo is the same but replacing $\boldsymbol{F}_{ele}$ with $\boldsymbol{F}^{\prime}_{ele}$.

%% file: sec/5_experiments.tex
\section{Experiments} \label{sec:experiments}
In this section, we comprehensively test performance of the proposed models and verify their superiority for practical RSR applications. We compare RoadBEV-mono with existing monocular depth estimation models, and RoadBEV-stereo with stereo matching methods. Ablation and comparison studies are implemented to investigate the influence of various parameters.

\subsection{Implementation Details} \label{implementation_details}

We utilize the following metrics to evaluate model performance: absolute error (abs. err.), root mean squared error (RMSE), ratio of grids with error greater than 0.5 cm (\textgreater0.5 cm), and frames per second (FPS) in inference.
We implement the models on PyTorch platform and train on RTX 3090 GPUs. Batch size is 8 for all experiments. Learning rate is set as 8e-4 for RoadBEV-mono, while 5e-4 for RoadBEV-stereo. The optimizer is AdamW with weight decay 1e-4. The OneCycle learning rate scheduler with linear decreasing strategy is used. The images are cropped to 960*528 to meet the size requirements. The RoadBEV-mono is trained for 50 epochs, while 40 for RoadBEV-stereo. Pre-trained weights on ImageNet is loaded. The compared models in Table \ref{tab:comparison_all} are also trained for the same epochs, with their default configurations in the provided codes.

\subsection{Performance and Comparison}
\begin{figure}[t]
\setlength{\abovecaptionskip}{0.3cm}
\setlength{\belowcaptionskip}{-0.2cm}
  \centering
   \includegraphics[width=1\linewidth]{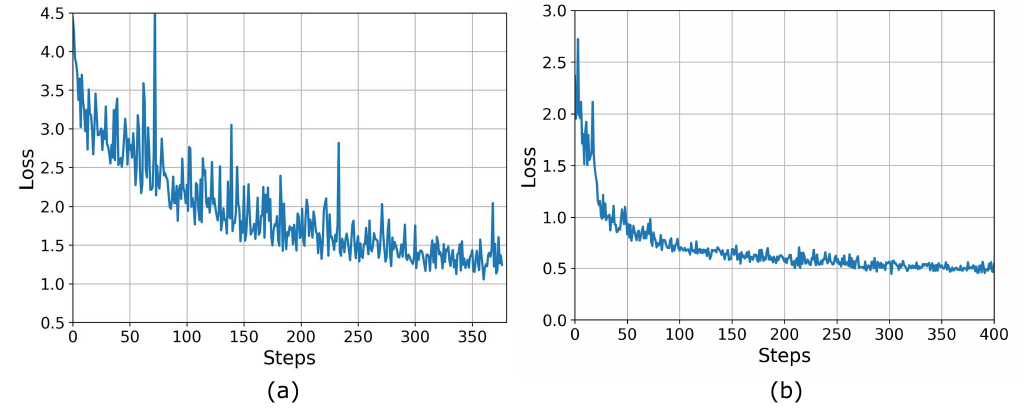}
   \vspace{-6pt}
   \caption{\my{Training losses of (a). RoadBEV-mono and (b). RoadBEV-stereo.}}
   \label{fig:loss}
\end{figure}

Fig. \ref{fig:loss} shows training losses of the proposed two models. They all reach convergence at the set training epochs, corroborating the effectiveness of model structures and loss functions. For the same GT labels and loss functions, the stereo-based model outperforms the monocular-based with lower loss value. The convergence of RoadBEV-stereo is more stable and faster. Comparing with the direct but inexplicable fitting in RoadBEV-mono, the introduction of stereo information significantly contributes more clues to learn road features in vertical direction.

As shown in Table \ref{tab:comparison_all}, we compare with depth estimation and stereo matching methods that ever achieved the SOTA performance on public datasets. Since the compared models finally provide depth in camera's coordinate, we convert them into BEV and generate elevation maps with the same style as GT labels.
For monocular-based road reconstruction, our model promotes the metrics with dominant advantages. \my{The absolute elevation error and RMSE reduces by 29.3\% and 25.8\% than the AdaBins, respectively.} The direct elevation estimation in BEV takes effect by uniformly extracting and aggregating features in the vertical direction. The error level at 1.83 cm captures most severe road unevenness that corrupts vehicle ride comfort, but not adequate for slighter road undulations. Moreover, the AdaBins achieves higher accuracy than the others using direct regression, verifying the necessity of classification on bins.

\begin{table}[t]
  \centering\addtolength{\tabcolsep}{-2.0pt}
  \caption{Performance comparison with monocular depth estimation and stereo matching methods. \textbf{Bold}: best.}
    \begin{tabular}{c|c||c|c|c|c}
    \hline
    \multicolumn{2}{c||}{\textbf{Method}} & \makecell[c]{Abs. err. \\ (cm)} & \makecell[c]{RMSE \\ (cm)} & \makecell[c]{\textgreater0.5 cm \\ (\%)} & FPS  \\
    \hline
    \multirow{5}{*}{\rotatebox{90}{Mono}} & LapDepth\cite{9316778} & 2.81 & 3.12 & 85.3 & 83.2\\
     & PixelFormer\cite{Agarwal_2023_WACV} & 2.65 & 2.86 & 82.0 & 43.0\\
     & iDisc\cite{piccinelli2023idisc} & 2.64 & 2.88 & 84.3 & 12.3 \\
     & AdaBins\cite{9578024} & 2.59 & 2.79 & 82.4 & 21.5 \\
     & RoadBEV(Ours) & \textbf{1.83} & \textbf{2.07} & \textbf{78.6} & 26.8 \\
    \hline
    \multirow{7}{*}{\rotatebox{90}{Stereo}} & IGEV-Stereo\cite{xu2023iterative} & 0.651 & 0.797 & 49.5 & 4.6 \\
     & PSMNet\cite{chang2018pyramid} & 0.654 & 0.785 & 50.1 & 11.6 \\
     & CFNet\cite{Shen_2021_CVPR}  & 0.647 & 0.760 & 50.8 & 6.8 \\
     & ACVNet\cite{xu2022attention} & 0.596 &  0.723 & 46.2 & 12.0 \\
     & GwcNet\cite{guo2019group} & 0.588 &  0.711 & 44.9 & 15.6 \\
     & DVANet\cite{zhao2024depth} & 0.546 &  0.685 & 40.9 & 8.7 \\
     & RoadBEV (Ours) & \textbf{0.503} & \textbf{0.609}  & \textbf{37.0} & 8.0\\
    \hline
    \end{tabular}%
  \label{tab:comparison_all}%
\end{table}%

Further, all the stereo-based models in Table \ref{tab:comparison_all} outperform the monocular-based with immense margins. Our RoadBEV-stereo also performs better than the others with noteworthy improvements. Although the absolute elevation error achieves 0.5 cm, only 37.0\% grids have error bigger than 0.5 cm, indicating that most estimations are concentrated within a small error range. The error level of 5.0 mm covers almost all road unevenness causing vertical vibration to vehicles. Our model has great potential in on-board RSR for aiding planning and control systems of UGVs. The proposed BEV cost volume and estimation head jointly function and efficiently recognize feature patterns in BEV. The constraints on search range simplify similarity measurement and identification with effectiveness. Comparing the architectures of RoadBEV-mono (Fig. \ref{fig:mono}) and RoadBEV-stereo (Fig. \ref{fig:stereo}), the principal promotion of RoadBEV-stereo lays in the utilization of extra information from another perspective (i.e., stereo image). The introduction of another view contributes to more than twice the performance enhancement. Therefore, stereo-based RSR is undoubtedly more promising and reliable than the monocular-based.

As for inference, our RoadBEV-mono achieves 26.8 FPS, which is promising for real-vehicle applications. The inference speed of RoadBEV-stereo is low due to the 3D convolutions requiring much computation. The inference speed is expected to be promoted in embedded deployment with special operation optimization. Despite that, since the camera has preview distance at about 7 m, the relative low inference speed is thereby compensated. Road elevation is still continuously available for the downstream planning and control units.

\begin{figure}[t]
  \centering
   \includegraphics[width=0.85\linewidth]{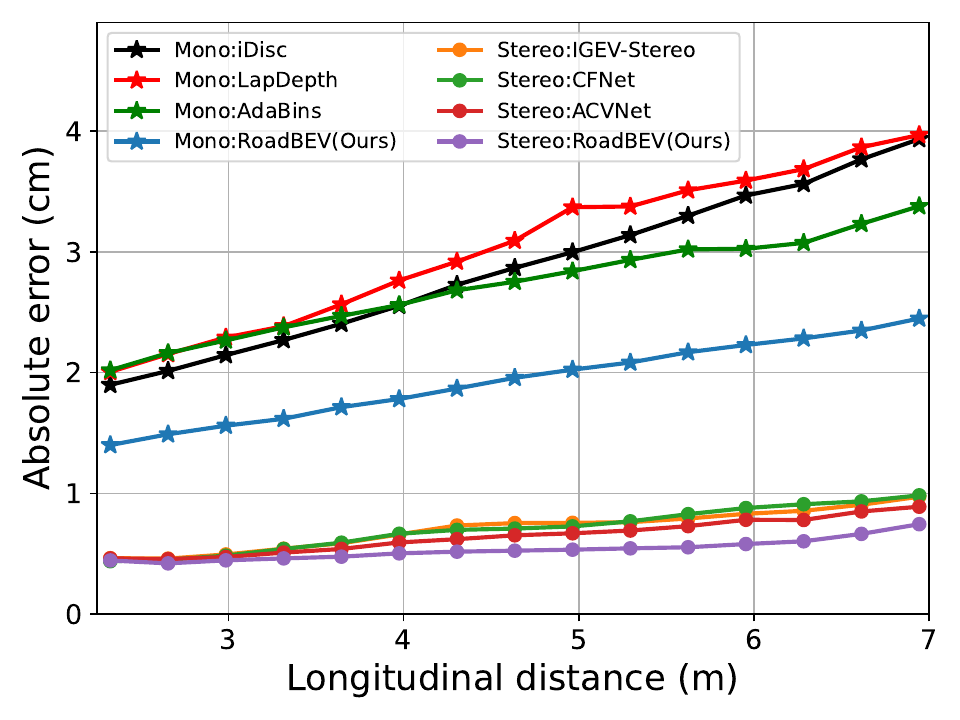}
   \vspace{-6pt}
   \caption{Comparison of distance-wise elevation errors with the SOTA models among both monocular-based and stereo-based.}
   \label{fig:comparison}
\end{figure}

For more insights, we further visualize the distance-wise absolute error, as shown in Fig. \ref{fig:comparison}. The whole grid is split into 15 segments along the longitudinal direction, with each segment representing a region of 33 cm long. Absolute elevation errors of grids inside the segments are averaged. Instead of a global but coarse metric, this analysis method digs into model performance and gives more comprehensive evaluation. Our RoadBEV-mono thoroughly outperforms the compared depth estimation models among the whole range with a significant margin. The superiority is more prominent at far distance comparing with LapDepth. As analyzed in Section \ref{RoadBEV-mono}. The AdaBins performs well as it defines the searching bins rather than direct fitting without effective constraints. 

Our RoadBEV-stereo also outperforms the compared stereo matching models, especially at far distance. However, the improvement is not as remarkable as that in the monocular. Potential reason is that the models already reach very high accuracy, while making further promotion is challenging. Noise in labels may also limit further enhancement. Despite that, the effectiveness of our model is firmly verified. 

\subsection{Visualization of Road Reconstruction} \label{sec:visualization}
\begin{figure*}[t]
  \centering
   \includegraphics[width=1\linewidth]{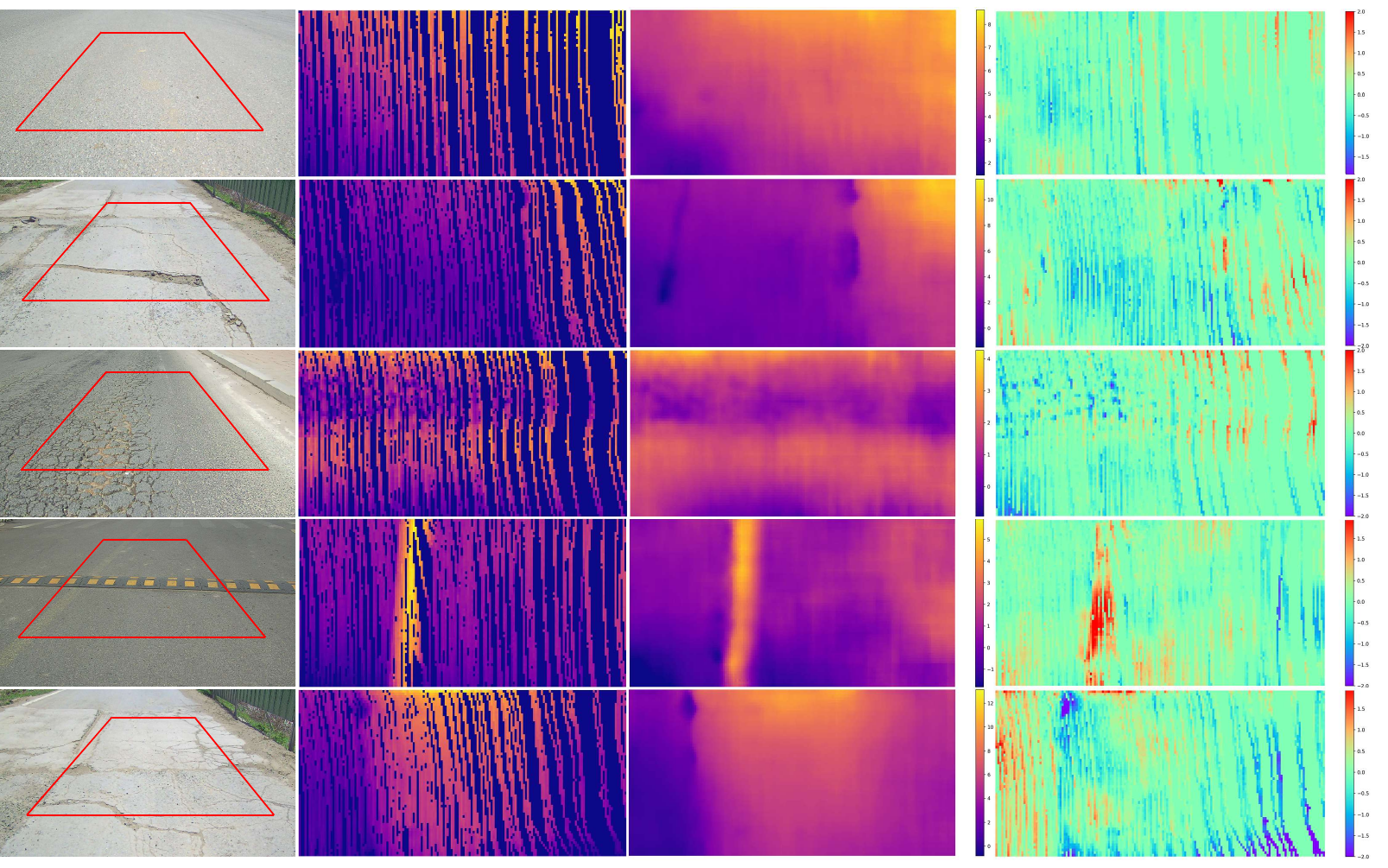}
   \vspace{-6pt}
   \caption{Visualization of road surface reconstructed by RoadBEV-mono. From left to right: RGB images, GT elevation maps, estimated elevation maps, and residual maps. The red bounding boxes indicate ROI. The residual map is calculated as $\boldsymbol{E}-\boldsymbol{\hat{E}}$.}
   \label{fig:visualization_mono}
\end{figure*}

The reconstructed road elevation maps by RoadBEV-mono are visualized in Fig. \ref{fig:visualization_mono}. Road surface structures and trends are accurately captured without holistic bias. Inferred map for the first sample, representing plane road surface without significant unevenness, shows stable elevations. This is crucial for downstream tasks as false detections may leads to unnecessary actions. The elongate crack in the second sample, as well as the abrupt bump at far distance, are recovered with high accuracy, as the residual map does not show prominent errors in the corresponding areas. The rut in the third sample is also caught, verifying its capability in recovering both global structures and local fine variations. The straight speed bump in the forth sample coincides well with GT. Although the amplitude error is relatively big in the middle part, it can still be clearly identified as a bump. The maximum absolute error can be bounded in 2.5 cm. Post-processing methods can be implemented to refine the bump's profile. 

\begin{figure*}[t]
  \centering
   \includegraphics[width=1\linewidth]{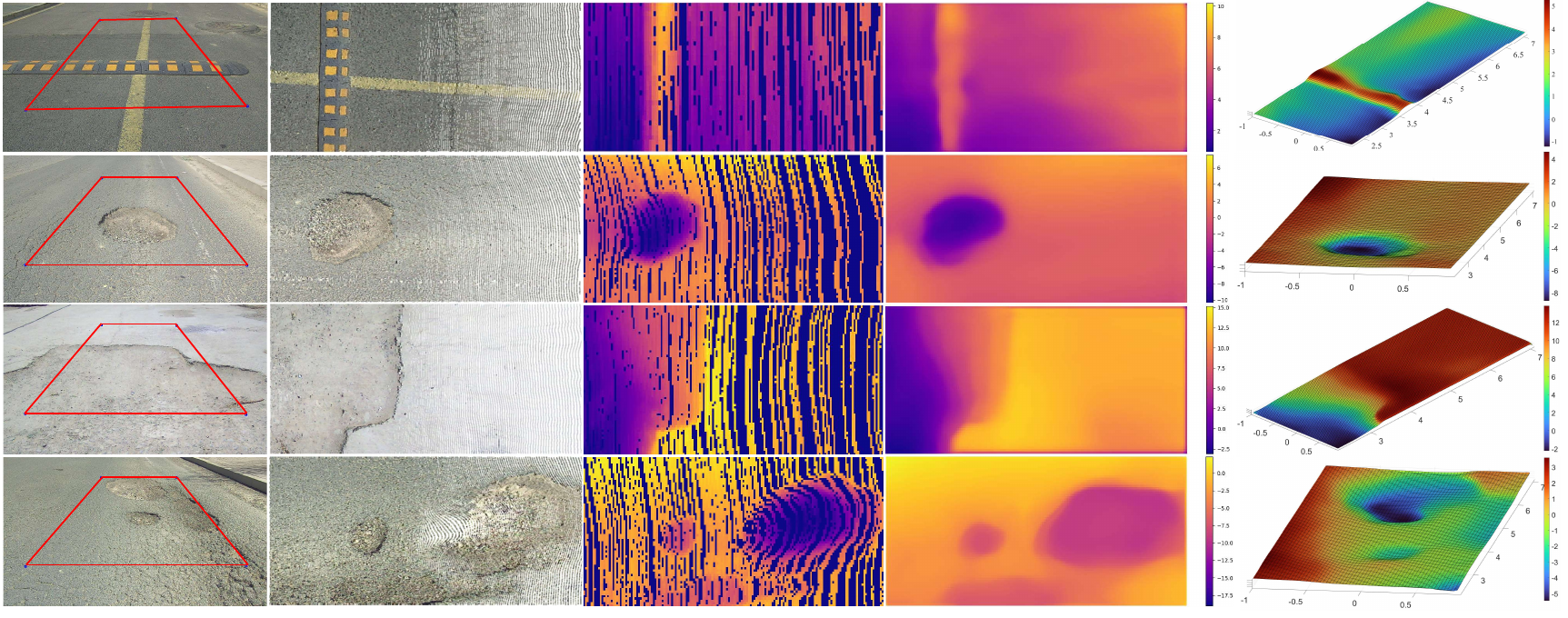}
   \vspace{-6pt}
   \caption{Visualization of road surface reconstructed by RoadBEV-stereo. From left to right: left images, recovered road point clouds with color in BEV, GT elevation maps, estimated elevation maps, and estimated road meshes.}
   \label{fig:visualization_stereo}
\end{figure*}

Fig. \ref{fig:visualization_stereo} shows the reconstruction results by RoadBEV-stereo. The second column shows the point clouds with color in BEV, which are obtained by GwcNet \cite{guo2019group}. The last column visualizes reconstructed 3D road meshes. Comparing with those in Fig. \ref{fig:visualization_mono}, the recovered elevation maps are smoother without cluttered noise or unexpected patterns. The RoadBEV-stereo is capable of preserving more detailed structures than RoadBEV-mono. The shapes and edges of potholes in the second and last samples are precisely recovered. As illustrated by the 3D meshes, the slopes from pothole edges to centers are also captured with exactness. In the third sample, the sharp steep between the cracked and even areas are clearly identifiable. All the presented samples demonstrate the capability of our method in handling complicated road patterns. 

\subsection{Ablation Studies for RoadBEV-mono} \label{ablation_mono}
We first explore the influence of class interval on model performance, as listed in Table \ref{tab:ablation_mono_class}. We set different class intervals between 0.4$\sim$2.0 cm, among which the metrics reach highest at 0.5 cm. Results also show that metric difference for class resolutions $\{0.5,0.8,1.0,1.6 \}$ is slight. Potential reason is that we utilize \emph{soft argmin} to regress elevation instead of picking the peak class. The weighted sum operation compensate accuracy decrease caused by discrete class interval. Over-high resolution at 0.4 cm results in $Nc$=100, which is too complicated to distinguish minute distinctness. Over-low resolution at 2.0 cm is too sparse to accurately represent the continuous elevation. Class interval is thereby set as 0.5 cm for both the proposed models.

\begin{table}[t]
  \centering
  \caption{Influence of class interval on performance of RoadBEV-mono. \textbf{Bold}: best.}
    \begin{tabular}{cc||c|c|c}
    \hline
    \makecell[c]{Class interval \\ (cm)} & \# classes & \makecell[c]{Abs. err. \\ (cm)} & \makecell[c]{RMSE \\ (cm)} & \makecell[c]{\textgreater0.5 cm \\ (\%)} \\
    \hline
      2.0 & 20 & 2.09 & 2.27 & 81.0 \\
      1.6 &  25 & 1.94 & 2.18 & 78.6 \\
      1.0 & 40 & 1.91 & 2.16 & 79.1 \\
      0.8 & 50 & 1.84 & 2.07 & 79.3 \\
      0.5 & 80 & \textbf{1.83} & \textbf{2.07} & \textbf{78.6} \\
      0.4 & 100 & 1.95 & 2.21 & 80.0 \\
    \hline
    \end{tabular}%
  \label{tab:ablation_mono_class}%
\end{table}%

We then conduct more comparison experiments to get instructive insights, as enumerated in Table \ref{tab:ablation_mono_all}. We probe the effects of different voxel vertical resolutions and feature extraction backbones. Both the metrics decrease when adopting EfficientNet-B3 as backbone. Sufficient feature extraction is essential for the subsequent estimation head. Similar with class resolution, the voxel resolution also reaches optimum at a medium value, i.e., 1.0 cm. For large vertical interval, sparse samplings on the search range are inadequate for retaining informative feature. For denser sampling at 0.5 cm, which equals the optimal class resolution, performance deteriorates instead. Smaller interval may introduce repeated sampling features especially at far distance owing to the perspective effect, where adjacent voxels are projected onto same pixel.

\begin{table}[t]
  \centering
  \caption{Ablation and comparison studies for RoadBEV-mono. \textbf{Bold}: best.}
    \begin{tabular}{ccc|cc||c|c}
    \hline
     \multicolumn{3}{c|}{{Voxel res. (cm)}} & \multicolumn{2}{c||}{{Backbone}} & \multirow{2}{*}{\makecell[c]{Abs err. \\ (cm)}} & \multirow{2}{*}{\makecell[c]{RMSE \\ (cm)}} \\
    \cline{1-5} \multicolumn{1}{c|}{0.5} & \multicolumn{1}{c|}{1.0} & \multicolumn{1}{c|}{2.0} & \multicolumn{1}{c|}{Eff-B3} & \multicolumn{1}{c||}{Eff-B6} &  &  \\
    \hline
    & $\checkmark$  &  & $\checkmark$ &  & 1.89 & 2.12 \\
    &   & $\checkmark$ & $\checkmark$ &  & 1.92 & 2.16 \\
    \hline
    $\checkmark$ &   &  &  & $\checkmark$ & 1.91 & 2.15  \\
    & $\checkmark$  &   &  & $\checkmark$ & \textbf{1.83} & \textbf{2.07} \\
    &  & $\checkmark$ &  & $\checkmark$ & 1.87 & 2.10 \\
   \hline
  \end{tabular}%
  \label{tab:ablation_mono_all}%
\end{table}%

\subsection{Ablation Studies for RoadBEV-stereo} \label{ablation_stereo}
The influence of volume construction, feature map resolution, and voxel interval are investigated. The first three experiments in Table \ref{tab:ablation_stereo} exploit point-wise subtraction to construct BEV volume instead of correlation. The corresponding metrics are slightly lower than the correlation volume.

\begin{table*}[t]
  \centering
  \caption{Ablation and comparison studies for RoadBEV-stereo. \textbf{Bold}: best.}
    \begin{tabular}{cc|ccc|cc||c|c|c}
    \hline
    \multicolumn{2}{c|}{{Feature res.}} & \multicolumn{3}{c|}{{Voxel res. (cm)}} & \multicolumn{2}{c||}{{Volume constr.}} & \multirow{2}{*}{\makecell[c]{Abs err. \\ (cm)}} & \multirow{2}{*}{\makecell[c]{RMSE \\ (cm)}} & \multirow{2}{*}{\makecell[c]{\textgreater0.5 cm \\ (\%)}} \\
    \cline{1-7}  \multicolumn{1}{c|}{1/4} & \multicolumn{1}{c|}{1/2} & \multicolumn{1}{c|}{0.5} & \multicolumn{1}{c|}{1.0} & \multicolumn{1}{c|}{2.0} & \multicolumn{1}{c|}{Subtr.} & \multicolumn{1}{c||}{Corr.} &  &  \\
    \hline
    $\checkmark$ &   &   & $\checkmark$ &  &  $\checkmark$ &  & 0.538 & 0.660 & 40.8  \\
    &  $\checkmark$ &   & $\checkmark$  &   & $\checkmark$ &  & 0.512 & 0.623 & 38.2 \\
    &  $\checkmark$ &   &  & $\checkmark$ & $\checkmark$ &  & 0.512 & 0.621 & 38.5  \\
   \hline
   $\checkmark$ &   & $\checkmark$  &  &  &   & $\checkmark$ & 0.550 & 0.683 & 41.3  \\
   $\checkmark$ &   &   & $\checkmark$  &  &  & $\checkmark$ & 0.536 & 0.660 & 40.2 \\
   $\checkmark$ &   &   &   & $\checkmark$ &  & $\checkmark$ & 0.535 & 0.655 & 40.4  \\
    \hline
    & $\checkmark$  &  $\checkmark$ &   &  &  & $\checkmark$ & 0.518 & 0.620 & 38.1 \\
    & $\checkmark$  &   & $\checkmark$  &  &  & $\checkmark$ & \textbf{0.503} & \textbf{0.609} & \textbf{37.0} \\
    & $\checkmark$  &   &   & $\checkmark$ &  & $\checkmark$ & 0.512 & 0.618 & 37.8 \\
   \hline
  \end{tabular}%
  \label{tab:ablation_stereo}%
\end{table*}%

Results in the last six experiments imply that for a certain feature map resolution, distinct voxel intervals have very slight influence on estimation performance. The estimation head is effective for both sparse and dense feature samplings along the search ranges. The effect of feature map resolution is more significant, where the models with $\frac{1}{2}$ feature resolution are superior. For instance, the absolute error reduces from 0.550 cm to 0.518 cm when voxel resolution is 0.5 cm. Phenomena above verify the rationality of analyzed mechanism in Section \ref{RoadBEV-stereo}. Stereo matching in BEV takes effect and performs better when high-resolution feature maps are given.

%% file: sec/limitation.tex
\section{Limitations and Prospects} \label{limitation}
We have fully explored the performance and applicability of the proposed two models. The reconstructed road surface elevation has immense potential in benefiting planning, control, and testing of autonomous vehicles. Nevertheless, challenges exist for practical on-board applications.

Visualizations in Fig. \ref{fig:comparison} indicate that the increasing trend of error w.r.t. longitudinal distance still exists for both the monocular and stereo-based. This is an inherent drawback of perspective camera, where texture and structure details are lost at far distance. Although the BEV paradigm directly reconstructs road surface from a top-down view, the features are still extracted from perspective images. BEV is a promising way to suppress this phenomenon, with further efforts and more advanced strategies.

In our implementations, we only adopt frames at current time. RoadBEV-stereo demonstrates that introducing other perspectives profoundly promotes model performance. Therefore, utilizing sequence images is expected to bring further enhancements. In view transformation, we project voxel centers onto image plane and index the corresponding pixel features. For more accurate feature query, fusing nearby pixel features with spatial cross attention deserve exploration \cite{li2022bevformer}.

We investigate model performance at the same horizontal resolution (i.e., 164*64 grids). Similar to stereo matching, the BEV volume can be first constructed at lower resolution and then integrally interpolated to full resolution, which effectively reduces computation. In this paper, we focus only on reconstructing road geometry structure, i.e., elevation. For future research, joint geometry and texture reconstruction can be explored, with recent technologies like NeRF \cite{mildenhall2021nerf,li2024ddn,Ling_2024_CVPR} and 3D Gaussian Splatting \cite{10.1145/3592433,feng2024rogs}.

We utilize our previous work, RSRD, to produce this algorithm and application prototype. Although typical road conditions are covered, the diversity is still inadequate, especially various corner cases. Efforts should be put to contribute more high-quality road surface data.

%% file: sec/6_conslusion.tex
\section{Conclusion}

In this paper, we reconstruct road surface elevation in Bird's Eye View for the first time. Two models named RoadBEV-mono and RoadBEV-stereo based on monocular and stereo images are proposed and analyzed, respectively. We reveal that monocular estimation and stereo matching in BEV have the same mechanism with that in perspective view, while improves by narrowing search range and digging features directly in elevation direction. Comprehensive experiments on real-world dataset validate the feasibility and superiority of proposed, BEV volume, estimation head, and parameter settings. \my{For monocular camera, the reconstruction error in BEV reduces by about 30\% than that in perspective view. Meanwhile, in BEV, the accuracy performance with stereo camera is more than three times of the monocular.} Insightful analysis and instructions about the models are provided. Our pioneering exploration also extends valuable references to further research and applications associated with BEV perception, 3D reconstruction and 3D detection.

%% file: New_IEEEtran_how-to.bbl
\begin{thebibliography}{10}
\providecommand{\url}[1]{#1}
\csname url@rmstyle\endcsname
\providecommand{\newblock}{\relax}
\providecommand{\bibinfo}[2]{#2}
\providecommand\BIBentrySTDinterwordspacing{\spaceskip=0pt\relax}
\providecommand\BIBentryALTinterwordstretchfactor{4}
\providecommand\BIBentryALTinterwordspacing{\spaceskip=\fontdimen2\font plus
\BIBentryALTinterwordstretchfactor\fontdimen3\font minus \fontdimen4\font\relax}
\providecommand\BIBforeignlanguage[2]{{%
\expandafter\ifx\csname l@#1\endcsname\relax
\typeout{** WARNING: IEEEtran.bst: No hyphenation pattern has been}%
\typeout{** loaded for the language `#1'. Using the pattern for}%
\typeout{** the default language instead.}%
\else
\language=\csname l@#1\endcsname
\fi
#2}}

\bibitem{8626459}
H.~Marzbani, H.~Khayyam, C.~N. TO, D.~V. Quoc, and R.~N. Jazar, ``Autonomous vehicles: Autodriver algorithm and vehicle dynamics,'' \emph{IEEE Transactions on Vehicular Technology}, vol.~68, no.~4, pp. 3201--3211, 2019.

\bibitem{ZHAO2024111019}
T.~Zhao, P.~Guo, and Y.~Wei, ``Road friction estimation based on vision for safe autonomous driving,'' \emph{Mechanical Systems and Signal Processing}, vol. 208, p. 111019, 2024.

\bibitem{10367760}
Z.~Yao, X.~Li, B.~Lang, and M.~C. Chuah, ``Goal-lbp: Goal-based local behavior guided trajectory prediction for autonomous driving,'' \emph{IEEE Transactions on Intelligent Transportation Systems}, pp. 1--10, 2023.

\bibitem{10101715}
T.~Zhao, J.~He, J.~Lv, D.~Min, and Y.~Wei, ``A comprehensive implementation of road surface classification for vehicle driving assistance: Dataset, models, and deployment,'' \emph{IEEE Transactions on Intelligent Transportation Systems}, vol.~24, no.~8, pp. 8361--8370, 2023.

\bibitem{10329453}
T.~Zhao, P.~Guo, J.~He, and Y.~Wei, ``A hierarchical scheme of road unevenness perception with lidar for autonomous driving comfort,'' \emph{IEEE Transactions on Intelligent Vehicles}, vol.~9, no.~1, pp. 2439--2448, 2024.

\bibitem{ZHAO2022108483}
T.~Zhao and Y.~Wei, ``A road surface image dataset with detailed annotations for driving assistance applications,'' \emph{Data in Brief}, vol.~43, p. 108483, 2022.

\bibitem{8324512}
S.~Li, G.~Zhang, X.~Lei, X.~Yu, H.~Qian, and Y.~Xu, ``Trajectory tracking control of a unicycle-type mobile robot with a new planning algorithm,'' in \emph{2017 IEEE International Conference on Robotics and Biomimetics (ROBIO)}, 2017, pp. 780--786.

\bibitem{9830854}
C.~Li, T.~Trinh, L.~Wang, C.~Liu, M.~Tomizuka, and W.~Zhan, ``Efficient game-theoretic planning with prediction heuristic for socially-compliant autonomous driving,'' \emph{IEEE Robotics and Automation Letters}, vol.~7, no.~4, pp. 10\,248--10\,255, 2022.

\bibitem{LIANG2022109197}
G.~Liang, T.~Zhao, Z.~Shangguan, N.~Li, M.~Wu, J.~Lyu, Y.~Du, and Y.~Wei, ``Experimental study of road identification by lstm with application to adaptive suspension damping control,'' \emph{Mechanical Systems and Signal Processing}, vol. 177, p. 109197, 2022.

\bibitem{ni2020road}
T.~Ni, W.~Li, D.~Zhao, and Z.~Kong, ``Road profile estimation using a 3d sensor and intelligent vehicle,'' \emph{Sensors}, vol.~20, no.~13, p. 3676, 2020.

\bibitem{weng2024big}
Y.~Weng, ``Big data and machine learning in defence,'' \emph{International Journal of Computer Science and Information Technology}, vol.~16, no.~2, pp. 25--35, 2024.

\bibitem{9050489}
L.~Wang, D.~Zhao, T.~Ni, and S.~Liu, ``Extraction of preview elevation information based on terrain mapping and trajectory prediction in real-time,'' \emph{IEEE Access}, vol.~8, pp. 76\,618--76\,631, 2020.

\bibitem{9691345}
L.~Sun, H.~Zhang, and W.~Yin, ``Pseudo-lidar-based road detection,'' \emph{IEEE Transactions on Circuits and Systems for Video Technology}, vol.~32, no.~8, pp. 5386--5398, 2022.

\bibitem{6662410}
G.-T. Michailidis, R.~Pajarola, and I.~Andreadis, ``High performance stereo system for dense 3-d reconstruction,'' \emph{IEEE Transactions on Circuits and Systems for Video Technology}, vol.~24, no.~6, pp. 929--941, 2014.

\bibitem{9025600}
R.~Fan, J.~Jiao, J.~Pan, H.~Huang, S.~Shen, and M.~Liu, ``Real-time dense stereo embedded in a uav for road inspection,'' in \emph{2019 IEEE/CVF Conference on Computer Vision and Pattern Recognition Workshops (CVPRW)}, 2019, pp. 535--543.

\bibitem{xin2024parameter}
Y.~Xin, S.~Luo, H.~Zhou, J.~Du, X.~Liu, Y.~Fan, Q.~Li, and Y.~Du, ``Parameter-efficient fine-tuning for pre-trained vision models: A survey,'' \emph{arXiv preprint arXiv:2402.02242}, 2024.

\bibitem{zhao2024depth}
T.~Zhao, M.~Ding, W.~Zhan, M.~Tomizuka, and Y.~Wei, ``Depth-aware volume attention for texture-less stereo matching,'' \emph{arXiv preprint arXiv:2402.08931}, 2024.

\bibitem{10321736}
H.~Li, C.~Sima, J.~Dai, W.~Wang, L.~Lu, H.~Wang, J.~Zeng, Z.~Li, J.~Yang, H.~Deng, H.~Tian, E.~Xie, J.~Xie, L.~Chen, T.~Li, Y.~Li, Y.~Gao, X.~Jia, S.~Liu, J.~Shi, D.~Lin, and Y.~Qiao, ``Delving into the devils of bird’s-eye-view perception: A review, evaluation and recipe,'' \emph{IEEE Transactions on Pattern Analysis and Machine Intelligence}, vol.~46, no.~4, pp. 2151--2170, 2024.

\bibitem{10438483}
J.~Wang, F.~Li, Y.~An, X.~Zhang, and H.~Sun, ``Towards robust lidar-camera fusion in bev space via mutual deformable attention and temporal aggregation,'' \emph{IEEE Transactions on Circuits and Systems for Video Technology}, pp. 1--1, 2024.

\bibitem{xin2024vmt}
Y.~Xin, J.~Du, Q.~Wang, Z.~Lin, and K.~Yan, ``Vmt-adapter: Parameter-efficient transfer learning for multi-task dense scene understanding,'' in \emph{Proceedings of the AAAI Conference on Artificial Intelligence}, vol.~38, no.~14, 2024, pp. 16\,085--16\,093.

\bibitem{chen2024taskclip}
H.~Chen, W.~Huang, Y.~Ni, S.~Yun, F.~Wen, H.~Latapie, and M.~Imani, ``Taskclip: Extend large vision-language model for task oriented object detection,'' \emph{arXiv preprint arXiv:2403.08108}, 2024.

\bibitem{6957961}
T.~Shen, G.~Schamp, and M.~Haddad, ``Stereo vision based road surface preview,'' in \emph{17th International IEEE Conference on Intelligent Transportation Systems (ITSC)}, 2014, pp. 1843--1849.

\bibitem{8500608}
B.~Li, Y.~Guo, J.~Zhou, Y.~Cai, J.~Xiao, and W.~Zeng, ``Lane detection and road surface reconstruction based on multiple vanishing point \& symposia,'' in \emph{2018 IEEE Intelligent Vehicles Symposium (IV)}, 2018, pp. 209--214.

\bibitem{8636338}
A.~Dhiman, H.-J. Chien, and R.~Klette, ``A multi-frame stereo vision-based road profiling technique for distress analysis,'' in \emph{2018 15th International Symposium on Pervasive Systems, Algorithms and Networks (I-SPAN)}, 2018, pp. 7--14.

\bibitem{7797253}
B.~Jia, J.~Chen, and K.~Zhang, ``Drivable road reconstruction for intelligent vehicles based on two-view geometry,'' \emph{IEEE Transactions on Industrial Electronics}, vol.~64, no.~5, pp. 3696--3706, 2017.

\bibitem{feng2020road}
Y.~Feng, R.~Zhang, and S.~Zhai, ``Road elevation map estimation based on affine transformation and stereo matching,'' in \emph{Journal of Physics: Conference Series}, vol. 1601, no.~6, 2020, p. 062015.

\bibitem{8794039}
D.~Li and T.~Furukawa, ``Global vision-based reconstruction of three-dimensional road surfaces using adaptive extended kalman filter,'' in \emph{2019 International Conference on Robotics and Automation (ICRA)}, 2019, pp. 3860--3866.

\bibitem{mei2024rome}
R.~Mei, W.~Sui, J.~Zhang, X.~Qin, G.~Wang, T.~Peng, T.~Chen, and C.~Yang, ``Rome: Towards large scale road surface reconstruction via mesh representation,'' \emph{IEEE Transactions on Intelligent Vehicles}, 2024.

\bibitem{wu2024emie}
W.~Wu, Q.~Wang, G.~Wang, J.~Wang, T.~Zhao, Y.~Liu, D.~Gao, Z.~Liu, and H.~Wang, ``Emie-map: Large-scale road surface reconstruction based on explicit mesh and implicit encoding,'' \emph{arXiv preprint arXiv:2403.11789}, 2024.

\bibitem{wang2023multi}
L.~Wang, X.~Zhang, Z.~Song, J.~Bi, G.~Zhang, H.~Wei, L.~Tang, L.~Yang, J.~Li, C.~Jia, \emph{et~al.}, ``Multi-modal 3d object detection in autonomous driving: A survey and taxonomy,'' \emph{IEEE Transactions on Intelligent Vehicles}, 2023.

\bibitem{song2024robustness}
Z.~Song, L.~Liu, F.~Jia, Y.~Luo, G.~Zhang, L.~Yang, L.~Wang, and C.~Jia, ``Robustness-aware 3d object detection in autonomous driving: A review and outlook,'' \emph{arXiv preprint arXiv:2401.06542}, 2024.

\bibitem{zhang2023dual}
X.~Zhang, L.~Wang, J.~Chen, C.~Fang, L.~Yang, Z.~Song, G.~Yang, Y.~Wang, X.~Zhang, and J.~Li, ``Dual radar: A multi-modal dataset with dual 4d radar for autononous driving,'' \emph{arXiv preprint arXiv:2310.07602}, 2023.

\bibitem{CAI2024102245}
Y.~Cai, H.~Che, B.~Pan, M.-F. Leung, C.~Liu, and S.~Wen, ``Projected cross-view learning for unbalanced incomplete multi-view clustering,'' \emph{Information Fusion}, vol. 105, p. 102245, 2024.

\bibitem{song2024graphbev}
Z.~Song, L.~Yang, S.~Xu, L.~Liu, D.~Xu, C.~Jia, F.~Jia, and L.~Wang, ``Graphbev: Towards robust bev feature alignment for multi-modal 3d object detection,'' \emph{arXiv preprint arXiv:2403.11848}, 2024.

\bibitem{10568349}
L.~Yang, X.~Zhang, J.~Yu, J.~Li, T.~Zhao, L.~Wang, Y.~Huang, C.~Zhang, H.~Wang, and Y.~Li, ``Monogae: Roadside monocular 3d object detection with ground-aware embeddings,'' \emph{IEEE Transactions on Intelligent Transportation Systems}, pp. 1--15, 2024.

\bibitem{yang2023bevheight}
L.~Yang, K.~Yu, T.~Tang, J.~Li, K.~Yuan, L.~Wang, X.~Zhang, and P.~Chen, ``Bevheight: A robust framework for vision-based roadside 3d object detection,'' in \emph{Proceedings of the IEEE/CVF Conference on Computer Vision and Pattern Recognition}, 2023, pp. 21\,611--21\,620.

\bibitem{yang2024sgv3d}
L.~Yang, X.~Zhang, J.~Li, L.~Wang, C.~Zhang, L.~Ju, Z.~Li, and Y.~Shen, ``Sgv3d: Towards scenario generalization for vision-based roadside 3d object detection,'' \emph{arXiv preprint arXiv:2401.16110}, 2024.

\bibitem{song2024robofusion}
Z.~Song, G.~Zhang, L.~Liu, L.~Yang, S.~Xu, C.~Jia, F.~Jia, and L.~Wang, ``Robofusion: Towards robust multi-modal 3d obiect detection via sam,'' \emph{arXiv preprint arXiv:2401.03907}, 2024.

\bibitem{huang2023tri}
Y.~Huang, W.~Zheng, Y.~Zhang, J.~Zhou, and J.~Lu, ``Tri-perspective view for vision-based 3d semantic occupancy prediction,'' in \emph{Proceedings of the IEEE/CVF conference on computer vision and pattern recognition}, 2023, pp. 9223--9232.

\bibitem{wang2022sti}
Y.~Wang, H.~Pan, J.~Zhu, Y.-H. Wu, X.~Zhan, K.~Jiang, and D.~Yang, ``Be-sti: Spatial-temporal integrated network for class-agnostic motion prediction with bidirectional enhancement,'' in \emph{Proceedings of the IEEE/CVF Conference on Computer Vision and Pattern Recognition}, 2022, pp. 17\,093--17\,102.

\bibitem{li2022hdmapnet}
Q.~Li, Y.~Wang, Y.~Wang, and H.~Zhao, ``Hdmapnet: An online hd map construction and evaluation framework,'' in \emph{2022 International Conference on Robotics and Automation (ICRA)}.\hskip 1em plus 0.5em minus 0.4em\relax IEEE, 2022, pp. 4628--4634.

\bibitem{li2022bevformer}
Z.~Li, W.~Wang, H.~Li, E.~Xie, C.~Sima, T.~Lu, Y.~Qiao, and J.~Dai, ``Bevformer: Learning bird’s-eye-view representation from multi-camera images via spatiotemporal transformers,'' in \emph{European conference on computer vision}.\hskip 1em plus 0.5em minus 0.4em\relax Springer, 2022, pp. 1--18.

\bibitem{philion2020lift}
J.~Philion and S.~Fidler, ``Lift, splat, shoot: Encoding images from arbitrary camera rigs by implicitly unprojecting to 3d,'' in \emph{Proceedings of the European Conference on Computer Vision}, 2020.

\bibitem{huang2022bevdet4d}
J.~Huang and G.~Huang, ``Bevdet4d: Exploit temporal cues in multi-camera 3d object detection,'' \emph{arXiv preprint arXiv:2203.17054}, 2022.

\bibitem{yang2023bevheight++}
L.~Yang, T.~Tang, J.~Li, P.~Chen, K.~Yuan, L.~Wang, Y.~Huang, X.~Zhang, and K.~Yu, ``Bevheight++: Toward robust visual centric 3d object detection,'' \emph{arXiv preprint arXiv:2309.16179}, 2023.

\bibitem{huang2021bevdet}
J.~Huang, G.~Huang, Z.~Zhu, and D.~Du, ``Bevdet: High-performance multi-camera 3d object detection in bird-eye-view,'' \emph{arXiv preprint arXiv:2112.11790}, 2021.

\bibitem{li2023bevdepth}
Y.~Li, Z.~Ge, G.~Yu, J.~Yang, Z.~Wang, Y.~Shi, J.~Sun, and Z.~Li, ``Bevdepth: Acquisition of reliable depth for multi-view 3d object detection,'' in \emph{Proceedings of the AAAI Conference on Artificial Intelligence}, vol.~37, no.~2, 2023, pp. 1477--1485.

\bibitem{li2022bevstereo}
Y.~Li, H.~Bao, Z.~Ge, J.~Yang, J.~Sun, and Z.~Li, ``Bevstereo: Enhancing depth estimation in multi-view 3d object detection with dynamic temporal stereo,'' \emph{arXiv preprint arXiv:2209.10248}, 2022.

\bibitem{zhao2024road}
T.~Zhao, Y.~Xie, M.~Ding, L.~Yang, M.~Tomizuka, and Y.~Wei, ``A road surface reconstruction dataset for autonomous driving,'' \emph{Scientific data}, vol.~11, no.~1, p. 459, 2024.

\bibitem{tan2019efficientnet}
M.~Tan and Q.~Le, ``Efficientnet: Rethinking model scaling for convolutional neural networks,'' in \emph{International conference on machine learning}.\hskip 1em plus 0.5em minus 0.4em\relax PMLR, 2019, pp. 6105--6114.

\bibitem{xie2022m}
E.~Xie, Z.~Yu, D.~Zhou, J.~Philion, A.~Anandkumar, S.~Fidler, P.~Luo, and J.~M. Alvarez, ``M2bev: Multi-camera joint 3d detection and segmentation with unified birds-eye view representation,'' \emph{arXiv preprint arXiv:2204.05088}, 2022.

\bibitem{9578024}
S.~Farooq~Bhat, I.~Alhashim, and P.~Wonka, ``Adabins: Depth estimation using adaptive bins,'' in \emph{2021 IEEE/CVF Conference on Computer Vision and Pattern Recognition (CVPR)}, 2021, pp. 4008--4017.

\bibitem{9316778}
M.~Song, S.~Lim, and W.~Kim, ``Monocular depth estimation using laplacian pyramid-based depth residuals,'' \emph{IEEE Transactions on Circuits and Systems for Video Technology}, vol.~31, no.~11, pp. 4381--4393, 2021.

\bibitem{Agarwal_2023_WACV}
A.~Agarwal and C.~Arora, ``Attention attention everywhere: Monocular depth prediction with skip attention,'' in \emph{Proceedings of the IEEE/CVF Winter Conference on Applications of Computer Vision (WACV)}, January 2023, pp. 5861--5870.

\bibitem{piccinelli2023idisc}
L.~Piccinelli, C.~Sakaridis, and F.~Yu, ``idisc: Internal discretization for monocular depth estimation,'' in \emph{IEEE Conference on Computer Vision and Pattern Recognition (CVPR)}, 2023.

\bibitem{xu2023iterative}
G.~Xu, X.~Wang, X.~Ding, and X.~Yang, ``Iterative geometry encoding volume for stereo matching,'' in \emph{Proceedings of the IEEE/CVF Conference on Computer Vision and Pattern Recognition}, 2023, pp. 21\,919--21\,928.

\bibitem{chang2018pyramid}
J.-R. Chang and Y.-S. Chen, ``Pyramid stereo matching network,'' in \emph{Proceedings of the IEEE Conference on Computer Vision and Pattern Recognition}, 2018, pp. 5410--5418.

\bibitem{Shen_2021_CVPR}
Z.~Shen, Y.~Dai, and Z.~Rao, ``Cfnet: Cascade and fused cost volume for robust stereo matching,'' in \emph{Proceedings of the IEEE/CVF Conference on Computer Vision and Pattern Recognition (CVPR)}, June 2021, pp. 13\,906--13\,915.

\bibitem{xu2022attention}
G.~Xu, J.~Cheng, P.~Guo, and X.~Yang, ``Attention concatenation volume for accurate and efficient stereo matching,'' in \emph{Proceedings of the IEEE/CVF Conference on Computer Vision and Pattern Recognition}, 2022, pp. 12\,981--12\,990.

\bibitem{guo2019group}
X.~Guo, K.~Yang, W.~Yang, X.~Wang, and H.~Li, ``Group-wise correlation stereo network,'' in \emph{Proceedings of the IEEE Conference on Computer Vision and Pattern Recognition}, 2019, pp. 3273--3282.

\bibitem{mildenhall2021nerf}
B.~Mildenhall, P.~P. Srinivasan, M.~Tancik, J.~T. Barron, R.~Ramamoorthi, and R.~Ng, ``Nerf: Representing scenes as neural radiance fields for view synthesis,'' \emph{Communications of the ACM}, vol.~65, no.~1, pp. 99--106, 2021.

\bibitem{li2024ddn}
M.~Li, J.~He, G.~Jiang, and H.~Wang, ``Ddn-slam: Real-time dense dynamic neural implicit slam with joint semantic encoding,'' \emph{arXiv preprint arXiv:2401.01545}, 2024.

\bibitem{Ling_2024_CVPR}
L.~Ling, Y.~Sheng, Z.~Tu, W.~Zhao, C.~Xin, K.~Wan, L.~Yu, Q.~Guo, Z.~Yu, Y.~Lu, X.~Li, X.~Sun, R.~Ashok, A.~Mukherjee, H.~Kang, X.~Kong, G.~Hua, T.~Zhang, B.~Benes, and A.~Bera, ``Dl3dv-10k: A large-scale scene dataset for deep learning-based 3d vision,'' in \emph{Proceedings of the IEEE/CVF Conference on Computer Vision and Pattern Recognition (CVPR)}, June 2024, pp. 22\,160--22\,169.

\bibitem{10.1145/3592433}
B.~Kerbl, G.~Kopanas, T.~Leimkuehler, and G.~Drettakis, ``3d gaussian splatting for real-time radiance field rendering,'' \emph{ACM Trans. Graph.}, vol.~42, no.~4, pp. 1--14, jul 2023.

\bibitem{feng2024rogs}
Z.~Feng, W.~Wu, and H.~Wang, ``Rogs: Large scale road surface reconstruction based on 2d gaussian splatting,'' \emph{arXiv preprint arXiv:2405.14342}, 2024.

\end{thebibliography}
